\documentclass{article}

\usepackage[preprint]{neurips_2025}

\usepackage[utf8]{inputenc} 
\usepackage[T1]{fontenc}    
\usepackage{hyperref}       
\usepackage{url}            
\usepackage{booktabs}       
\usepackage{amsfonts}       
\usepackage{nicefrac}       
\usepackage{microtype}      
\usepackage{xcolor}         

\usepackage{multirow}
\usepackage{multicol}
\usepackage{makecell}
\usepackage{enumitem}
\usepackage{balance}
\usepackage{graphicx}
\usepackage{subfigure}
\usepackage{subcaption}
\usepackage{caption}
\usepackage{amsmath}
\usepackage{amsthm}
\usepackage{amssymb}
\usepackage{amsfonts}
\usepackage{wrapfig}
\usepackage{lipsum}
\usepackage{arydshln}

\title{Scaling over Scaling: Exploring Test-Time Scaling Plateau in Large Reasoning Models}

\author{%
Jian Wang, ~~Boyan Zhu, ~~Chak Tou Leong, ~~Yongqi Li\thanks{Corresponding author.}, ~~Wenjie Li \\
Department of Computing, The Hong Kong Polytechnic University \\
\texttt{jian51.wang@polyu.edu.hk} ~~\texttt{\{boyan.zhu,chak-tou.leong\}@connect.polyu.hk} \\
\texttt{liyongqi0@gmail.com} ~~\texttt{cswjli@comp.polyu.edu.hk}
}

\begin{document}

\maketitle

\begin{abstract}
Large reasoning models (LRMs) have exhibited the capacity of enhancing reasoning performance via internal test-time scaling.
Building upon this, a promising direction is to further scale test-time compute to unlock even greater reasoning capabilities. 
However, as we push these scaling boundaries, systematically understanding the practical limits and achieving optimal resource allocation becomes a critical challenge. 
In this paper, we investigate the \textit{scaling plateau} of test-time scaling and introduce the Test-Time Scaling Performance Model (TTSPM). We theoretically analyze two fundamental paradigms for such extended scaling, parallel scaling and sequential scaling, from a probabilistic modeling perspective. 
Our primary contribution is the derivation of the saturation point on the scaling budget for both strategies, identifying thresholds beyond which additional computation yields diminishing returns. Remarkably, despite their distinct mechanisms, both paradigms converge to a unified mathematical structure in their upper bounds.
We empirically validate our theoretical findings on challenging reasoning benchmarks, including AIME, MATH-500, and GPQA, demonstrating the practical utility of these bounds for test-time resource allocation. 
We hope that this work provides insights into the cost-benefit trade-offs of test-time scaling, guiding the development of more resource-efficient inference strategies for large reasoning models.

\end{abstract}

\section{Introduction}
\label{sec:intro}

Large language models (LLMs) have demonstrated impressive capabilities across a wide range of reasoning tasks. Their performance can be further enhanced by increasing compute at inference time (or test time), a technique commonly referred to as \textit{test-time scaling} (TTS)~\citep{zhang2025and}. Recent advancements have enabled LLMs to generate longer and more coherent Chain-of-Thought (CoT) reasoning traces, as seen in models like OpenAI's o1~\citep{openai2024o1}, o3~\citep{openai2025o3}, and DeepSeek-R1~\citep{guo2025deepseek}. This approach, known as \textit{internal scaling}, extends the model's deliberation within a single forward pass, significantly advancing LLMs into a new class of large reasoning models (LRMs).

Building upon the foundation of internally scaled reasoning, we naturally pose the next question: \textit{Can these already powerful LRMs be further enhanced by allocating even more compute during inference?} This inquiry motivates our exploration of test-time scaling over internally scaled LRMs, a concept we refer to as \textbf{scaling over scaling}. However, as we consider increasing test-time compute, such as by generating additional solution candidates or performing more rounds of iterative refinement, it becomes essential to ask: Is such scaling always beneficial? In other words, \textit{does a point of diminishing returns exist, beyond which additional computational effort yields negligible performance gains?} Identifying and characterizing this \textbf{scaling plateau} is critical for preventing resource over-allocation, especially in most TTS strategies where the absence of supervised feedback complicates efficient compute usage.

To address the above questions, we introduce the \textbf{T}est-\textbf{T}ime \textbf{S}caling \textbf{P}erformance \textbf{M}odel (\textbf{TTSPM}), a theoretical probabilistic model that reimagines test-time scaling performance through the lens of probabilistic modeling (as detailed in Section~\ref{sec:method}).
TTSPM begins with two fundamental TTS paradigms: 1) parallel scaling \citep{wang2023self}, where multiple reasoning paths and solutions are generated independently; and 2) sequential scaling \citep{tian2025think}, where a solution is iteratively refined round by round.
Their performance gains with increasing compute often exhibit a characteristic saturation curve. 
We develop a general probabilistic model that captures this behavior, positing that the probability of success with the scaling budget $N$ approaches a maximum achievable performance $F_{\text{max}}$ as $N$ grows. Based on this probabilistic model, we analyze the marginal performance gain per additional generation, which is referred to as a scaling unit. We then formally define the \textit{scaling plateau} as the point where this marginal gain is no longer higher than a predefined small threshold $\epsilon$ as increasing scaling units.
Using this definition, we derive a unified saturation point for the scaling budget, which signifies the critical point beyond which further scaling is deemed inefficient.

We empirically validate the practical utility of our theoretical bounds on challenging mathematical reasoning and PhD-level science question answering benchmarks, including AIME \citep{maa2024aime}, MATH-500 \citep{lightman2024let}, and GPQA \citep{rein2024gpqa}. Our experiments (see Section~\ref{sec:experiments}) demonstrate that the derived bounds effectively predict the onset of the scaling plateau, aligning with observed performance saturation. These findings offer valuable insights. First, LRMs can indeed be further improved via simple, verifier-free test-time scaling. Second, this improvement is not limitless and exhibits a predictable saturation point. Our TTSPM provides a principled method to estimate this point, enabling more efficient resource allocation by avoiding unnecessary token consumption.

Overall, the contributions of this paper are as follows: 
1) We propose TTSPM, a theoretical probabilistic model that effectively characterizes test-time scaling plateau in large reasoning models. 
2) With TTSPM, we derive a general upper bound for scaling units. It is applicable to both parallel and sequential scaling strategies, revealing a consistent mathematical structure underlying their saturation behavior. 
3) We outline how this theoretical bound can be empirically validated and discuss its practical implications for optimizing test-time compute, thereby guiding the development of more resource-efficient inference scaling strategies for large reasoning models.

\section{Related Work}
\label{sec:related_work}

The quest to enhance the reasoning abilities of large language models (LLMs) beyond standard pre-training and fine-tuning has led to significant interest in techniques that leverage increased computation during inference, commonly referred to as test-time scaling (TTS) \citep{zhang2025and,wu2025inference}. These methods aim to improve performance on complex tasks by allocating more resources for ``thinking'' or exploration before generating final answers, without modifying the model's parameters. Test-time scaling strategies can be broadly categorized based on \textit{what} aspect of the generation process is scaled and \textit{how} the scaling is implemented \citep{zhang2025and}.

\paragraph{Parallel Test-Time Scaling.}

Parallel scaling methods explore multiple reasoning paths or solutions concurrently. Self-Consistency (SC) \citep{wang2023self} is a prominent example, where multiple reasoning paths are sampled independently using Chain-of-Thought (CoT), and the final answer is determined by majority voting over the outcomes. This approach improves robustness against occasional reasoning errors in single paths. Best-of-N (BoN) sampling \citep{brown2024large} also generates multiple candidate outputs and selects the best one based on a scoring mechanism, often using the model's own likelihood scores or an external verifier \citep{lightman2024let}. Tree of Thoughts (ToT) \citep{yao2023tree} explores reasoning as a tree search, where multiple reasoning paths are explored in parallel at each step. While parallel methods effectively broaden the search space, they often involve redundant computations as paths are generated independently without collaboration. Majority voting in SC can also be misled if multiple paths converge to the same incorrect answer. Recent work has also explored adaptive parallel reasoning \citep{pan2025learning}, but explicit theoretical mechanisms remain less studied.

\paragraph{Sequential Test-Time Scaling.}

Sequential scaling methods enhance reasoning by generating intermediate steps or thoughts in a serial manner. The seminal work on Chain-of-Thought (CoT) prompting \citep{wei2022chain} demonstrated that prompting LLMs to produce step-by-step reasoning traces significantly improves performance on arithmetic, commonsense, and symbolic reasoning tasks. Building upon this, various techniques have explored iterative refinement and self-correction. Self-Refine \citep{madaan2023self} enables models to iteratively refine their outputs based on self-generated feedback. Similarly, methods like \citep{li2025dancing} employ stepwise natural language self-critique to enhance reasoning. Other approaches focus on extending the length or depth of sequential thought. For example, \cite{tian2025think} proposed Multi-round Thinking, scaling the number of thinking rounds at test time. \citep{yan2025inftythink} suggested breaking length limits in long-context reasoning. While effective in guiding the model, purely sequential approaches can suffer from high latency, potential context length overflow, and the risk of error propagation, where an early mistake derails the entire reasoning process \citep{marjanovic2025deepseek, ye2025limo}.

\begin{wrapfigure}{r}{0.5\textwidth}
  \centering
  \includegraphics[width=0.48\textwidth]{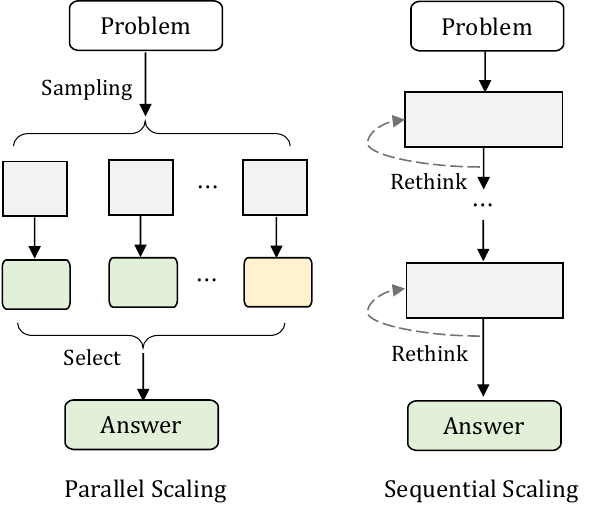}
  \caption{Conceptual illustration of parallel scaling vs. sequential scaling strategy.}
  \label{fig:example}
  \vspace{-10pt}
\end{wrapfigure}

\paragraph{Hybrid and Other Scaling Strategies.}

Recognizing the limitations of purely sequential or parallel methods, hybrid approaches attempt to combine their strengths. Some methods integrate verification steps within the generation process \citep{lightman2024let, chen2025sets, shi2025heimdall}. Reinforcement learning has also been employed to control the reasoning process or optimize test-time compute allocation \citep{aggarwal2025l1, qu2025optimizing, guo2025deepseek}. 
For example, \cite{li2025reasoning} proposed aligning reasoning with logic units. \cite{jin2025two} investigated multi-agent collaborative reasoning.
Recent work has also explored adaptive or hybrid approaches that try to combine the benefits of both parallel and sequential methods or dynamically allocate compute \citep{pan2025learning, tan2025adaptive, liu2025metascale, qu2025optimizing, paliotta2025thinking, mei2025a1}. 
They primarily investigated the role of verifiers or reinforcement learning to guide the scaling process \citep{aggarwal2025l1, shi2025heimdall, setlur2025scaling, chen2025sets}.

However, a fundamental question that remains underexplored is the inherent limit of performance gain from these test-time scaling strategies, particularly for verifier-free approaches. While it is intuitive that more computation should lead to better performance, the relationship is unlikely to be linear indefinitely. Understanding when these scaling efforts hit a point of diminishing returns—a \textit{scaling plateau}—is critical for efficient resource utilization. 
Our work attempts to address this gap by theoretically modeling this trend and deriving explicit upper bounds (saturation points) on the computational budget for both parallel and sequential scaling, beyond which further scaling offers minimal performance gains. This provides a principled basis for optimizing test-time compute allocation, complementing existing empirical explorations of TTS efficiency.

\section{Test-Time Scaling Performance Model}
\label{sec:method}

Building on the premise that large reasoning models (LRMs) can be further improved by scaling test-time computation, but that such improvements are not limitless, we aim to build a theoretical \textbf{T}est-\textbf{T}ime \textbf{S}caling \textbf{P}erformance \textbf{M}odel (\textbf{TTSPM}) to quantify reasoning performance gain varying by scaling budget.
TTSPM captures the saturation behavior common to various test-time scaling strategies, despite their differing underlying mechanisms. 
From this model, we derive a general upper bound (saturation point), identifying the scaling budget beyond which further scaling efforts yield diminishing returns below a practical threshold.

\subsection{Probabilistic Modeling of Test-Time Scaling}
\label{ssec:prob_model}

Despite test-time scaling strategies differing in their fine-grained dynamics, a common macro-level phenomenon is widely observed: as more generations per problem are invested at test time, the performance tends to improve but eventually saturates, approaching a practical maximum. 
To capture this overarching saturation behavior in a general manner, we start from two fundamental scaling paradigms: \textit{parallel scaling} and \textit{sequential scaling}.

\paragraph{Parallel Scaling: Sampling to Selection.}
Given an input problem $q$ and a large reasoning model $\mathcal{M}$, parallel scaling consists of two crucial steps (as shown in Figure~\ref{fig:example}): generating $N$ candidate answers $\{a_i\}_{i=1}^N$ (i.e., sampling) and then applying a majority voting~\citep{wang2023self} mechanism or taking a verifier~\citep{wang2024math} to select the best answer based on the scores of these candidates (i.e., selection strategy, denoted as $\mathcal V$). 
Each answer $a_i$ is either correct ($c_i=1$) or incorrect ($c_i=0$ ). 
The probability of any single candidate answer $a_i$ being deemed correct is given by:
\begin{equation}
    p_{\text{sample}} = P(c_i=1 | q, \mathcal{M}, \mathcal{V}).
\end{equation}

Let $K$ be the number of correct answers among these $N$ candidates. Since each answer is independently assessed with probability $p_{\text{sample}}$ of being correct, $K$ follows a Binomial distribution.
We are interested in the event that at least one answer is correct, i.e., $K \geq 1$. The probability is given by:
\begin{equation}
    P(K \geq 1) = 1 - P(K=0),
\end{equation}
where $P(K=0)$ is the probability that all $N$ answers are incorrect. Due to the independence of parallel sampling, we have:
\begin{equation}
    P(K=0) = (P(c_{i}=0))^N = (1-p_{\text{sample}})^{N}.
\end{equation}
Thus, the probability of finding at least one correct answer is:
\begin{equation}
    P(K \geq 1) = 1 - (1-p_{\text{sample}})^{N}.
\end{equation}

\paragraph{Sequential Scaling: Round-by-round Rethinking.}

Sequential scaling is concluded as asking the model to iteratively rethink its previous answer round by round (see Figure~\ref{fig:example}).
We model this as a discrete-time Markov process with two states: the current answer (denoted as $s_0$) and the correct answer (denoted as $s_1$). State $s_1$ is an absorbing state, meaning once a correct answer is reached, the process stops and remains correct.
Given the model $\mathcal{M}$ and the input problem $q$, let $p_{\text{rethink}}$ be the probability of transitioning from an any state $s_0$ to the correct state $s_1$ in a single round:
\begin{equation}
    p_{\text{rethink}} = P(s_t = s_1 | s_{t-1} = s_0, q, \mathcal{M}).
\end{equation}
This probability $p_{\text{rethink}}$ quantifies the model's intrinsic capability to self-correct or improve its answer within a single iteration. Consequently, the probability of remaining in state $s_0$ (i.e., failing to correct the answer in one round) is $1-p_{\text{rethink}}$.
Let $K$ be the random variable representing the number of rethinking rounds required to first reach the correct state $S_1$. For the process to first reach $S_1$ at round $k$ (where $k \geq 1$), it must have remained in $S_0$ for the preceding $k-1$ rounds and then transitioned to $S_1$ at the $k$-th round. The probability of this sequence of events is $P(K=k) = (1-p_{\text{rethink}})^{k-1}p_{\text{rethink}}$. This indicates that $K$ follows a geometric distribution with success parameter $p_{\text{rethink}}$.

We are primarily interested in the cumulative probability of achieving a correct answer within a maximum of $N$ rethinking rounds, denoted $P(K \leq N)$. This is the sum of probabilities of first reaching $S_1$ in any round from $1$ to $N$:
\begin{equation}
    \begin{aligned}
        P(K \leq N) &= \sum_{k=1}^{N} P(K=k) = \sum_{k=1}^{N}(1-p_{\text{rethink}})^{k-1}p_{\text{rethink}}.
    \end{aligned}
\end{equation}
This sum represents the first $N$ terms of a geometric series. The first term is $p_{\text{rethink}}$ (for $k=1$) and the common ratio is $(1-p_{\text{rethink}})$. The sum evaluates to:
\begin{equation}
\begin{aligned}
    P(K \leq N) &= p_{\text{rethink}} \frac{1 - (1-p_{\text{rethink}})^N}{1 - (1-p_{\text{rethink}})} \\
    &= p_{\text{rethink}} \frac{1 - (1-p_{\text{rethink}})^N}{p_{\text{rethink}}} \\
    &= 1 - (1-p_{\text{rethink}})^N.
\end{aligned}
\end{equation}

\label{eq:sequential_prob_success_N_rounds}
This equation provides the probability of attaining a correct answer via sequential rethinking within $N$ rounds. This formulation is structurally analogous to the success probability in parallel scaling, $1 - (1-p_{\text{sample}})^N$, differing primarily in the interpretation of the base success probability.

\subsection{Performance Function to Scaling Plateau}
\label{ssec:performance}

\paragraph{Performance Function.}
Without loss of generality, we let $p_{x}$ denote the probability of attaining a correct answer, either by parallel scaling ($p_{\text{sample}}$) or by sequential scaling ($p_{\text{rethink}}$). We define $N$ as the \textit{scaling budget}, which refers to the number of generations per problem, with each generation denoted as a scaling unit. We use $F(N)$ to represent the model performance (e.g., Hit@$k$ or Pass@$k$), given by:
\begin{equation}
F(N) = F_{\text{max}} \cdot (1 - (1-p_x)^N),
\label{eq:perf_model}
\end{equation}
where $F_{\text{max}}$ ($0 < F_{\text{max}} \leq 1$) represents the theoretical maximum performance achievable by the model $\mathcal{M}$ on the specific task, even with an unbounded scaling budget. It is determined by the inherent capabilities and potential limitations of the model and training for the task at hand.

\paragraph{Marginal Performance Gain.}
To precisely identify the onset of diminishing returns, we introduce the marginal performance gain, $\Delta F(N)$, which is the additional performance improvement obtained by increasing the scaling budget from $N$ to $N+1$ units. This is given by:
\begin{align}
\Delta F(N) &= F(N+1) - F(N) \nonumber \\
&= F_{\text{max}} \cdot (1 - (1-p_x)^{N+1}) - F_{\text{max}} \cdot (1 - (1-p_x)^N) \nonumber \\
&= F_{\text{max}} \cdot [(1-p_x)^N - (1-p_x)^{N+1}] \nonumber \\
&= F_{\text{max}} \cdot (1-p_x)^N [1 - (1-p_x)] \nonumber \\
&= F_{\text{max}} \cdot p_x \cdot (1-p_x)^N.
\label{eq:delta_perf}
\end{align}
Eq.~(\ref{eq:delta_perf}) shows that $\Delta F(N)$ decreases exponentially as $N$ increases (since $0 < 1-p_x < 1$). This exponential decay signifies that each successive scaling unit contributes progressively less to the overall performance improvement than its predecessor.

\paragraph{Derivation of Scaling Plateau.}
We define the \textit{scaling plateau} as the operational region where this marginal performance gain, $\Delta F(N)$, becomes practically insignificant. Specifically, we posit that further scaling is inefficient if $\Delta F(N)$ falls below a predefined small positive threshold $\epsilon$. This threshold $\epsilon$ represents the minimum performance improvement deemed worthwhile for the cost of an additional scaling unit.
Our objective is to identify an upper bound for the scaling budget, $N_{\text{upper}}$, such that for any $N \geq N_{\text{upper}}$, the marginal gain $\Delta F(N)$ is consistently less than the chosen threshold $\epsilon$. This condition is formally expressed as:
\begin{equation}
F_{\text{max}} \cdot p_x \cdot (1-p_x)^N < \epsilon.
\label{eq:plateau_condition}
\end{equation}
To solve for $N$, we rearrange the terms:
\begin{equation}
(1-p_x)^N < \frac{\epsilon}{F_{\text{max}} \cdot p_x}.
\label{eq:rearranged_condition}
\end{equation}
For this inequality to have a meaningful solution for $N \geq 1$ and for the subsequent logarithmic operations to be well-defined, we require $0 < \frac{\epsilon}{F_{\text{max}} \cdot p_x} < 1$. This condition implies that $\epsilon < F_{\text{max}} \cdot p_x$, meaning the chosen threshold for negligible gain must be smaller than the initial marginal gain provided by the first scaling unit (i.e., $\Delta F(0) = F_{\text{max}} \cdot p_x$). If this condition is not met (i.e., $\epsilon \geq F_{\max} \cdot p_x$), it signifies that even the very first scaling unit does not provide a sufficient performance boost, and thus $N_{\text{upper}}$ could be considered to be 0 or 1.

Assuming $\epsilon < F_{\max} \cdot p_x$, we take the natural logarithm of both sides of Eq.~(\ref{eq:rearranged_condition}):
\begin{equation}
N \ln(1-p_x) < \ln\left(\frac{\epsilon}{F_{\max} \cdot p_x}\right).\end{equation}
Given that $0 < p_x < 1$, it follows that $0 < 1-p_x < 1$, which makes $\ln(1-p_x)$ a negative value. Therefore, when dividing by $\ln(1-p_x)$, the direction of the inequality sign is reversed:
\begin{equation}
N > \frac{\ln\left(\frac{\epsilon}{F_{\max} \cdot p_x}\right)}{\ln(1-p_x)}.
\label{eq:N_critical_raw}
\end{equation}
According to the right-hand side of Eq.~(\ref{eq:N_critical_raw}), the smallest integer $N$ that satisfies this inequality represents the point at which the marginal gain consistently drops below $\epsilon$. We define this as the unified upper bound (or saturation point), $N^{*}$, which is given by:
\begin{equation}
  N^{*} = \left\lceil \frac{\ln\left(\frac{\epsilon}{F_{\max} \cdot p_x}\right)}{\ln(1-p_x)} \right\rceil.
\label{eq:N_upper_general}
\end{equation}
This $N^{*}$ signifies the scaling budget beyond which scaling additional units is expected to yield performance improvements below the threshold $\epsilon$, effectively marking the scaling plateau.

\paragraph{Insights into the Scaling Plateau.}

The core finding of this paper is that both parallel and sequential test-time scaling strategies, despite their operational differences, exhibit a performance saturation phenomenon—the \textit{scaling plateau}—that can be characterized by a unified mathematical structure, as induced in Eq.~(\ref{eq:N_upper_general}). This suggests a fundamental principle at play: as more computational units ($N$) are invested at test time, the probability of encountering an unsolved problem that can be successfully addressed by the next unit diminishes. The exponential decay term $(1-p_x)^N$ in our marginal gain formula (Eq.~(\ref{eq:delta_perf})) captures this essence. The derived saturation point, $N^{*}$, provides a quantitative tool to identify when the expected marginal gain drops below a practical threshold $\epsilon$, signaling that further scaling is unlikely to be cost-effective.

\subsection{Practical Considerations and Parameter Estimation}
\label{ssec:discussion_practical}

Applying the scaling budget bound in practice requires estimating the model parameters of the maximum performance $F_{\max}$, effective success probability $p_x$ (either $p_{\text{sample}}$ or $p_{\text{rethink}}$), and setting a suitable gain threshold $\epsilon$.
We observe that: 1) $F_{\max}$ is challenging to determine precisely. It can be empirically estimated by running the LRM with a very large $N$ on a representative validation set and observing the performance asymptote. Alternatively, it could be set based on known human performance levels or theoretical limits for a given task. The choice of $F_{\max}$ will influence $N^{*}$; a higher $F_{\max}$ (if achievable) might justify a larger $N^{*}$.
2) $p_x$ is specific to the scaling strategy and the model-task combination. For parallel scaling ($p_{\text{sample}}$), it can be estimated from the initial rate of performance increase with $N$ or by analyzing the success rate of individual samples on a validation set. For sequential scaling ($p_{\text{rethink}}$), it reflects the probability of successful correction/improvement per step, which can also be estimated from validation data. These parameters might not be constant across all problem instances, which is a simplification in our current model.
3) $\epsilon$ is a user-defined hyperparameter reflecting the acceptable trade-off between performance gain and computational cost. A smaller $\epsilon$ will lead to a larger $N^{*}$, indicating a willingness to invest more compute for smaller gains. Its choice depends on the application context, available resources, and latency requirements.
\section{Experiments and Analysis}
\label{sec:experiments}

\subsection{Experimental Setup}
\label{ssec:exp_setup}

\paragraph{Benchmarks and Backbone Models.} 

We experiment on multiple representative reasoning benchmarks. \textbf{AIME 2024}~\cite{maa2024aime} and \textbf{AIME 2025}\footnote{\url{https://huggingface.co/datasets/math-ai/aime25}.}: Each contains 30 pre-Olympiad level problems from the American Invitational Mathematics Examination, designed to test advanced mathematical reasoning. \textbf{MATH-500}~\cite{hendrycks2measuring}: A challenging subset of the MATH dataset with 500 high school competition problems across algebra, geometry, and so on. \textbf{GPQA}~\cite{rein2024gpqa}: A science-domain question answering benchmark involving PhD-level science questions.
We employ \texttt{DeepSeek-R1-Distill-Qwen-1.5B}~\cite{guo2025deepseek} and \texttt{DeepSeek-R1-Distill-Qwen-7B}~\cite{guo2025deepseek} as our large reasoning models for investigation.

\paragraph{Test-Time Scaling Setup.} 
We implement parallel scaling based on Self-Consistency~\cite{wang2023self}, which generates $N$ candidate answers and applies a majority voting to select the final answer. We implement sequential scaling by asking the model to rethink its previous answer and obtain the final answer round by round, similar to Multi-round Thinking~\cite{tian2025think}. 
To assess the test-time scaling performance of different methods, we vary the number of generations per problem (i.e., $N$) from 1 to 32.
For all models, we use a sampling temperature of 0.6 and a top-$p$ of 0.95, following the suggested parameter settings~\cite{guo2025deepseek}. We set the maximum length to 32,000, allowing the model to perform sufficient reasoning. All experiments are conducted in one NVIDIA A6000 server with 8 GPUs.
We report the metrics of Accuracy (Acc.) and Hit@$N$. Accuracy measures the percentage of problems solved correctly, while Hit@$N$ evaluates the proportion of problems for which at least one correct solution is found among $N$ generated outputs.

\begin{table}[t!]
\caption{Comparison of different test-time scaling methods on several representative reasoning benchmarks. ``R1-1.5B'' and ``R1-7B'' represent \texttt{DeepSeek-R1-Distill-Qwen-1.5B} and \texttt{DeepSeek-R1-Distill-Qwen-7B}, respectively. ``Seq.'' and ``Par.'' are short for \textit{sequential} scaling and \textit{parallel} scaling, respectively.}
\vspace{5pt}
\label{tab:main_results}
\centering
\resizebox{0.98\textwidth}{!}{
\begin{tabular}{l l cc cc cc cc}
\toprule
\multirow{2}{*}{\textbf{Model}}  & \multirow{2}{*}{\textbf{Method}} & \multicolumn{2}{c}{\textbf{AIME 2024}} &  \multicolumn{2}{c}{\textbf{AIME 2025}} &  \multicolumn{2}{c}{\textbf{MATH-500}}  & \multicolumn{2}{c}{\textbf{GPQA}} \\
\cmidrule{3-10}
  & & Acc. & Hit@$N$ &  Acc. & Hit@$N$  & Acc. & Hit@$N$ & Acc. & Hit@$N$ \\
\midrule
\multirow{7}{*}{R1-1.5B}  & Vanilla &  30.0 &  30.0 & 23.3 & 23.3  &  74.2 & 74.2 &  21.3 & 21.3  \\
  &  Seq. ($N=4$) &  36.7 & 43.3 & 23.3 & 30.0 &  78.4 & 86.6  & 34.8  & 44.4 \\
  &  Seq. ($N=8$) &  26.7 & 46.7 & 26.7 & 33.3  &  76.9 & 88.2  &  35.4 & 51.0 \\
  &  Seq. ($N=32$) &  30.0 & 70.0 & 26.7 & 43.3 &  80.6 & 90.6 &  35.4 & 72.2  \\
\cmidrule{2-10}
  &  Par.  ($N=4$)  & 30.0 & 56.7 & 30.0 & 33.3 &  79.2 & 86.4 &  39.4 & 71.7  \\
  &  Par.  ($N=8$)  &  46.7 & 60.0 & 33.3 & 40.0 & 81.6 & 89.2 & \textbf{39.9} & 85.4 \\
  &  Par.  ($N=32$) & \textbf{53.3} & \textbf{80.0}  &  \textbf{40.0} & \textbf{56.7}  &  \textbf{83.6} & \textbf{91.8} & 39.4 & \textbf{96.5} \\
\midrule
\multirow{7}{*}{R1-7B}  & Vanilla & 56.7 & 56.7 & 40.0 & 40.0  & 82.0 & 82.0 &  41.9 & 41.9   \\
  &  Seq. ($N=4$) & 63.3 & 70.0  & 53.3 &  60.0 & 85.8 &  88.2 & 41.4 & 55.6 \\
  &  Seq. ($N=8$) & 63.3 & 76.7  & 40.0 &  60.0 &  85.0 & 88.2 & 44.9 & 55.6 \\
  &  Seq. ($N=32$) & 73.3 & 76.7  & 43.3 & 60.0  &  86.4 & 89.6  & 46.5 & 58.6 \\
\cmidrule{2-10}
  &  Par.  ($N=4$)  & 73.3 & 76.7  &  53.3 & 60.0  &  85.8 & 88.6 & 55.1 & 75.8 \\
  &  Par.  ($N=8$)  & \textbf{76.7} & 80.0 &  \textbf{60.0} & 60.0   & 85.4 &  90.4 & 54.5 & 85.5  \\
  &  Par.  ($N=32$)  & \textbf{76.7} &  \textbf{86.7} &  53.3 & \textbf{70.0}  &  \textbf{86.8} & \textbf{92.2} & \textbf{56.6} & \textbf{90.4}  \\
\bottomrule
\end{tabular}}
\end{table}

\subsection{Main Results of Test-Time Scaling}

As shown in Table~\ref{tab:main_results}, both sequential and parallel test-time scaling strategies significantly improve the performance of large reasoning models across all benchmarks. First, we observe that test-time scaling consistently enhances model performance regardless of model size. For the smaller 1.5B model, parallel scaling with $N=32$ improves accuracy by up to 23.3 percentage points on AIME 2024 (from 30.0\% to 53.3\%) and percentage points on GPQA (from 21.3\% to 39.4\%). Similarly, the larger 7B model shows substantial gains, with improvements of up to 20.0 percentage points on AIME 2024 (from 56.7\% to 76.7\%) and 14.7 percentage points on GPQA (from 41.9\% to 56.6\%).
Second, parallel scaling consistently outperforms sequential scaling across all benchmarks and model sizes. This performance gap is particularly pronounced for the 1.5B model on AIME 2024, where parallel scaling with $N=32$ achieves 53.3\% accuracy compared to 30.0\% with sequential scaling at the same budget. The superior performance of parallel scaling can be attributed to its ability to explore a more diverse solution space through independent sampling, whereas sequential scaling may suffer from error propagation or local optima in the refinement process.

\begin{figure}[t!]
    \centering
   \includegraphics[width=1.0\textwidth]{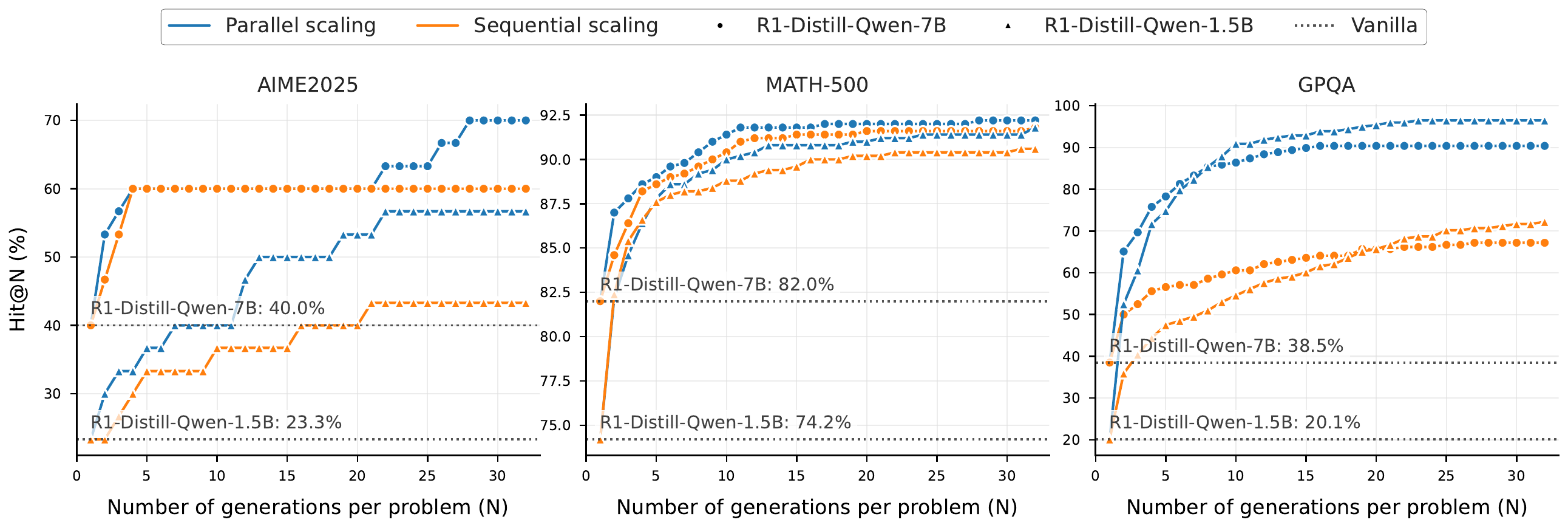}
    \caption{Scaling curves across three benchmarks (including AIME 2025, MATH-500, and GPQA), illustrating how Hit@$N$ varies with the number of generations per problem ($N$) under different scaling strategies (parallel vs. sequential) and model sizes.}
    \label{fig:scaling_curve}
\end{figure}

\begin{figure}[t!]
    \centering
   \includegraphics[width=1.0\textwidth]{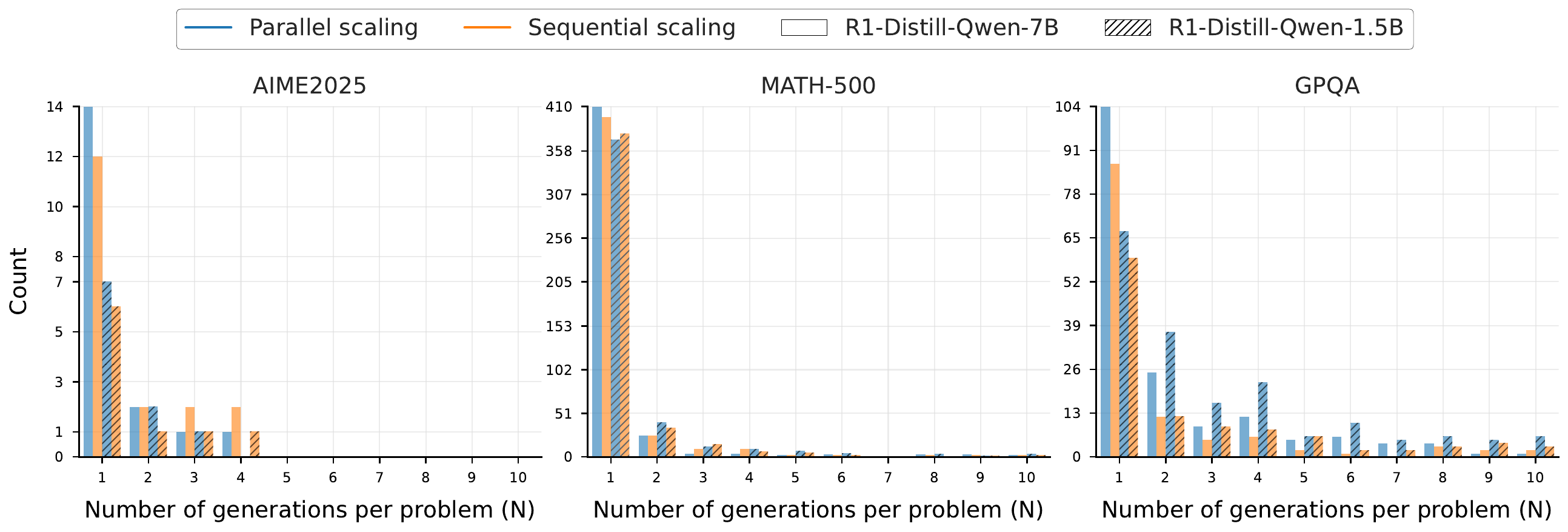}
    \caption{Statistics of the number of generations per problem ($N$) required to reach saturation points, comparing different scaling strategies (parallel vs. sequential) and model sizes.}
    \vspace{-10pt}
    \label{fig:stats_scaling}
\end{figure}

\subsection{Analysis of the Scaling Plateau}

Figure~\ref{fig:scaling_curve} shows the scaling curves across different benchmarks, model sizes, and scaling strategies, exhibiting a consistent pattern of initial rapid improvement followed by a plateau, confirming our theoretical model. We conclude that several important insights emerge from these curves:
1) \textbf{The scaling plateau is a universal phenomenon observed across all experimental settings}. For both the 1.5B and 7B models, performance improvements become increasingly marginal as the number of generations increases, eventually reaching a point where additional scaling yields negligible gains. This empirical observation strongly validates our theoretical derivation of the scaling plateau.
2) \textbf{Through test-time scaling, a smaller model can approach and sometimes even surpass the performance of a larger model without scaling}. On MATH-500, the 1.5B model with parallel scaling at $N=32$ achieves 83.6\% accuracy, exceeding the 7B model's vanilla performance of 82.0\%. This demonstrates that test-time scaling can be a cost-effective alternative to model size scaling, offering a practical approach to enhance reasoning capabilities without the computational burden of training larger models.
3) \textbf{The distribution discrepancy of different benchmark results in distinct scaling characteristics}. GPQA shows the steepest initial improvement, particularly for parallel scaling, suggesting that this benchmark benefits most from diverse solution exploration. In contrast, AIME2025 exhibits more gradual improvements, indicating that the problems in this benchmark may require more focused refinement rather than broad exploration.
Further accuracy scaling curves presented in Appendix~\ref{app:results_acc_curve} demonstrate similar insights.

Figure~\ref{fig:stats_scaling} outlines the statistics of the number of generations per problem required to reach saturation points. We observe that the distribution of optimal scaling budgets varies significantly across benchmarks and scaling strategies. For AIME 2025, both models show a strong concentration at $N=1$, indicating that many problems either can be solved in the first attempt or remain challenging regardless of additional scaling. In contrast, GPQA exhibits a more dispersed distribution, particularly for the 1.5B model with sequential scaling, suggesting that problems in this benchmark benefit from varying degrees of computational investment.

\subsection{Verification of the Scaling Plateau}
\label{ssec:verification_scaling}
To validate our theoretical model, TTSPM, we compare the predicted scaling plateau with the empirically observed optimal scaling budget for each problem in the MATH-500 dataset. To achieve a theoretical prediction of the scaling plateau of $N$, we should first estimate the value of the probability $p_x$, so that we can employ Eq. (\ref{eq:N_upper_general}) to calculate the saturation point of the scaled units. We randomly split the MATH-500 dataset into a ``validation set'' and a ``test set'' in a proportion of 8:2. We then utilize the ``validation set'' to estimate the unknown entry $\frac{\epsilon}{F_{\max}}$. With this, we can calculate the predicted scaling plateau on the test set. Further details are presented in Appendix~\ref{sec:appendix_exp}.

Figure~\ref{fig:verification} presents the correlation between the theoretically predicted and the practically observed scaling saturation points in the MATH-500 dataset. The strong positive correlations observed in all different models and scaling strategies provide compelling evidence for the predictive power of our theoretical model. Parallel scaling shows particularly strong correlations, with Pearson's correlation coefficient $r$ = 0.768 for the 1.5B model and $r$ = 0.803 for the 7B model.
These correlations indicate that our theoretical model can effectively predict the point at which additional scaling becomes inefficient, allowing us to make informed decisions about computational resource allocation. The stronger correlation observed for parallel scaling suggests that our model is particularly well-suited for predicting the scaling behavior of independent sampling approaches.
Interestingly, we observe that the 7B model with parallel scaling (Figure~\ref{fig:scatter_plot-7b_para}) shows the highest correlation, indicating that larger, more stable models may be more predictable in their scaling behavior. This finding has important implications for resource allocation in large-scale deployments, where accurate prediction of scaling benefits can lead to significant computational savings.

\begin{figure}[t]
    \centering
    \subfigure[Sequential Scaling (1.5B)]{
        \includegraphics[width=0.45\linewidth]{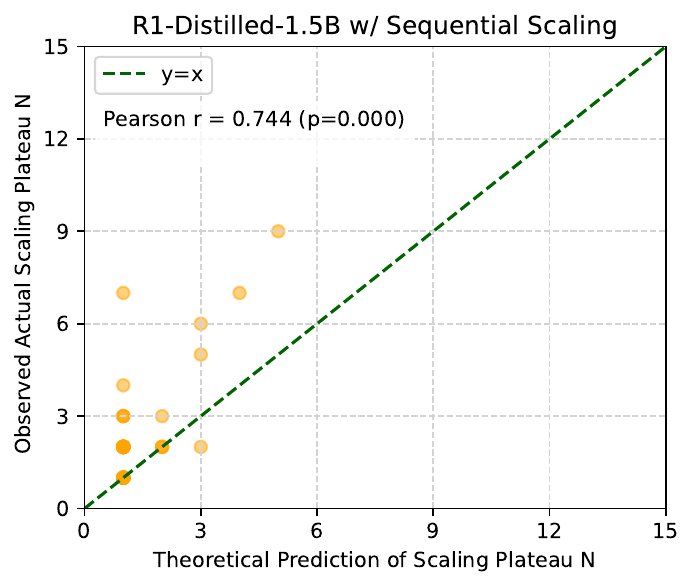}
        \label{fig:scatter_plot-1.5b_seq}
    }
    \subfigure[Parallel Scaling (1.5B)]{
        \includegraphics[width=0.45\linewidth]{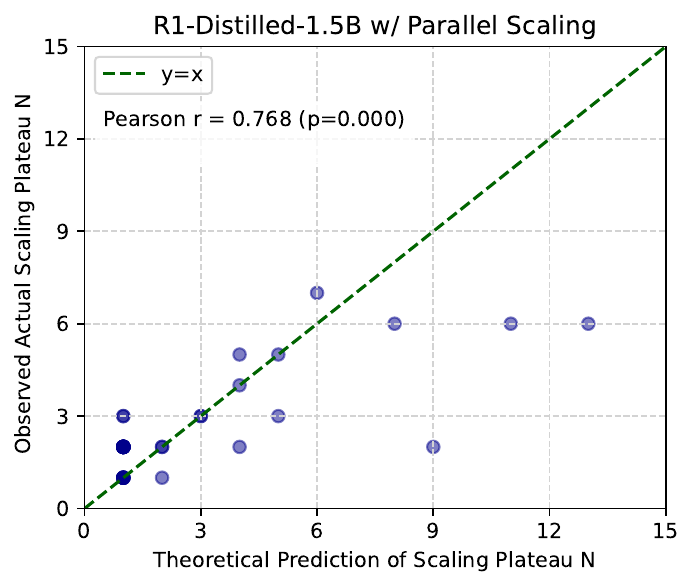}
        \label{fig:scatter_plot-1.5b_para}
    }
    \subfigure[Sequential Scaling (7B)]{
        \includegraphics[width=0.45\linewidth]{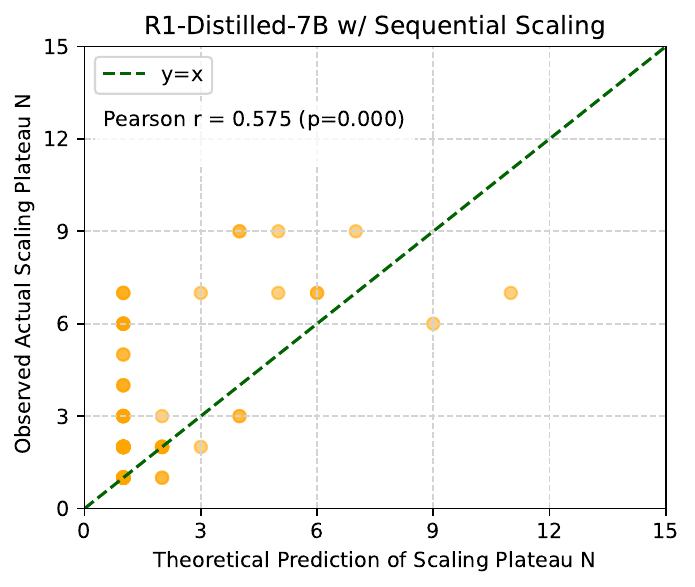}
        \label{fig:scatter_plot-7b_seq}
    }
    \subfigure[Parallel Scaling (7B)]{
        \includegraphics[width=0.45\linewidth]{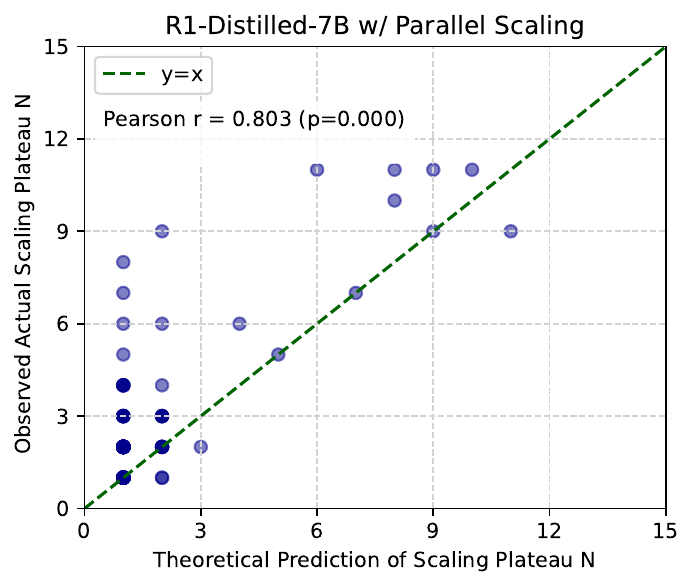}
        \label{fig:scatter_plot-7b_para}
    }
    \caption{Correlation between the theoretically predicted and empirically observed scaling plateau points (i.e., the number of generations per problem $N$) across different scaling strategies (sequential vs. parallel) and model sizes (1.5B vs. 7B).}
    \vspace{-5pt}
    \label{fig:verification}
\end{figure}

\subsection{Discussion on Limitations}
\label{ssec:discussion_limitations}

Our theoretical model, while providing valuable insights, relies on several simplifying assumptions: 1) \textbf{Independence of Scaling Units.} The derivation $1-(1-p_x)^N$ assumes that the ``success events'' associated with each scaling unit are conditionally independent given the current state. While this is a common modeling assumption, complex dependencies might exist, especially in sequential scaling, where the outcome of one step heavily influences the next.
2) \textbf{Definition of ``Success'' and $p_x$.} The precise definition and empirical estimation of an ``effective success probability'' $p_x$ can be nuanced. It encapsulates not just the model's ability to generate a correct step/sample but also any implicit selection or aggregation mechanism (e.g., majority voting in self-consistency, or the criteria for a successful rethink).
3) \textbf{Verifier-Free Assumption.} Our primary focus is on scaling strategies that do not rely on external verifiers. The dynamics might change if a strong verifier is available to prune or guide the search, potentially altering the $p_x$ values or even the form of the performance curve.

\section{Conclusion and Future Work}

In this paper, we investigate the \textit{scaling plateau} phenomenon in test-time computation for large reasoning models, developing a unified theoretical test-time scaling performance model (TTSPM) that characterizes the point at which additional computational units yield diminishing returns. 
Our key contribution is the derivation of a general upper bound (saturation point) with identical mathematical structure for both parallel and sequential scaling strategies, suggesting a fundamental principle governs test-time scaling limits regardless of mechanism. 
Experiments on challenging reasoning benchmarks validate our theoretical bounds, enabling practitioners to make informed decisions about cost-benefit trade-offs in test-time scaling. 
In the future, we aim to enhance our method in several key aspects. Currently, our TTSPM is task- and model-specific, requiring careful instantiation and parameter estimation for each new setting. To address this limitation, we plan to adapt TTSPM to support rapid parameter estimation across diverse scenarios. 
Additionally, we intend to explore hybrid scaling strategies, develop dynamic parameter estimation techniques, and extend the core principles of our approach to a broader range of reasoning tasks.





{
\small
\bibliography{reference}
\bibliographystyle{plain}
}

\newpage
\appendix

\section{Experimental Details}
\label{sec:appendix_exp}

Our theoretical model for the scaling saturation point, $N^{*}$, relies on the effective probability of success for a single scaling unit, $p(x)$, which varies across problems. We estimate this probability for each problem in both validation and test sets using experimental data from up to $N=32$ generations per problem.

\begin{itemize}[leftmargin=*]
    \item \textbf{For Parallel Scaling}: The effective probability $p(x)$ corresponds to $p_{\text{sample}}(x)$, the probability that a single candidate answer for problem $x$ is correct. We estimate this by:
    $$ \hat{p}_{\text{sample}}(x) = \frac{\text{Number of correct answers for problem } x \text{ in 32 samples}}{32} $$ 
    
    \item \textbf{For Sequential Scaling}: The effective probability $p(x)$ corresponds to $p_{\text{rethink}}(x)$, the probability of transitioning from any state to the correct state in a single rethinking round. For a problem first solved at round $k_x \leq 32$, we use the maximum likelihood estimate for a geometric distribution:
    $$ \hat{p}_{\text{rethink}}(x) = \begin{cases} 
    k_x/32 & \text{if problem } x \text{ is first solved at round } k_x \leq 32 \\ 
    1e-5 & \text{if problem } x \text{ is not solved within 32 rounds} 
    \end{cases} $$ 
    This approach aligns with our theoretical model, where $p(x)$ represents the single-step success probability in the Markov process. For problems not solved within 32 rounds, we assign a small non-zero probability to avoid mathematical issues in subsequent logarithmic calculations.
\end{itemize}

These estimations are performed for every problem in both the validation and test sets, providing the problem-specific parameters needed for our scaling plateau predictions.

\section{Probabilistic Modeling}
\label{app:appendix_prob}

This appendix provides supplementary mathematical details for the derivations presented in Section~\ref{sec:method}.

\subsection{Detailed Derivation of Marginal Performance Gain}
\label{app:delta_perf_detail}

Recall the unified performance model from Eq. (~\ref{eq:perf_model}):
\begin{equation}
F(N) = F_{\max} \cdot (1 - (1-p_x)^N) \tag{Eq. (\ref{eq:perf_model}) revisited}
\end{equation}
The marginal performance gain, $\Delta F(N)$, is defined as the difference in performance when increasing the computational budget from $N$ to $N+1$ units:
\begin{equation}
\Delta F(N) = F(N+1) - F(N)
\end{equation}
Substituting the performance model:
\begin{align}
\Delta F(N) &= \left[ F_{\max} \cdot (1 - (1-p_x)^{N+1}) \right] - \left[ F_{\max} \cdot (1 - (1-p_x)^N) \right] \\
&= F_{\max} \left[ (1 - (1-p_x)^{N+1}) - (1 - (1-p_x)^N) \right] \\
&= F_{\max} \left[ 1 - (1-p_x)^{N+1} - 1 + (1-p_x)^N \right] \\
&= F_{\max} \left[ (1-p_x)^N - (1-p_x)^{N+1} \right]
\end{align}
We can factor out $(1-p_x)^N$ from the terms in the bracket:
\begin{align}
\Delta F(N) &= F_{\max} \cdot (1-p_x)^N \left[ 1 - (1-p_x) \right] \\
&= F_{\max} \cdot (1-p_x)^N \left[ 1 - 1 + p_x \right] \\
&= F_{\max} \cdot p_x \cdot (1-p_x)^N 
\end{align}
This is Eq. (~\ref{eq:delta_perf}) in the main text.

\subsection{Detailed Derivation of the Inequality for \textit{N}}
\label{app:N_inequality_detail}

We start from the condition for the scaling plateau (Eq (~\ref{eq:plateau_condition})):
\begin{equation}
F_{\max} \cdot p_x \cdot (1-p_x)^N < \epsilon \tag{\ref{eq:plateau_condition} revisited}
\end{equation}
To solve for $N$, we first isolate the term $(1-p_x)^N$. Assuming $F_{\max} > 0$ and $p_x > 0$, we can divide both sides by $F_{\max} \cdot p_x$ without changing the inequality direction (as this product is positive):
\begin{equation}
(1-p_x)^N < \frac{\epsilon}{F_{\max} \cdot p_x}
\end{equation}
This is Equation~\ref{eq:rearranged_condition} in the main text.

For this inequality to be meaningful in the context of finding a positive $N$ where performance gain diminishes, we require the right-hand side to be less than 1 (since $(1-p_x)^N$ will be less than 1 for $N \geq 1$ and $0 < p_x < 1$). If $\frac{\epsilon}{F_{\max} \cdot p_x} \geq 1$, it means that even for $N=0$ (or $N=1$, depending on interpretation), the marginal gain $F_{\max} \cdot p_x$ is already less than or equal to $\epsilon$, implying the plateau is reached immediately. Thus, we assume $0 < \frac{\epsilon}{F_{\max} \cdot p_x} < 1$, which implies $\epsilon < F_{\max} \cdot p_x$.

Taking the natural logarithm of both sides:
\begin{equation}
\ln((1-p_x)^N) < \ln\left(\frac{\epsilon}{F_{\max} \cdot p_x}\right)
\end{equation}
Using the logarithm property $\ln(a^b) = b \ln(a)$:
\begin{equation}
N \ln(1-p_x) < \ln\left(\frac{\epsilon}{F_{\max} \cdot p_x}\right)
\end{equation}
Since $0 < p_x < 1$, it follows that $0 < 1-p_x < 1$. The natural logarithm of a number between 0 and 1 is negative, so $\ln(1-p_x) < 0$. When dividing an inequality by a negative number, the direction of the inequality sign must be reversed:
\begin{equation}
N > \frac{\ln\left(\frac{\epsilon}{F_{\max} \cdot p_x}\right)}{\ln(1-p_x)}
\end{equation}
This is Eq. (~\ref{eq:N_critical_raw}) in the main text. The unified upper bound $N^{*}$ is then defined as the smallest integer satisfying this condition.

\subsection{Condition for the Existence of a Non-trivial Scaling}
\label{app:N_upper_condition}

As mentioned in Section~\ref{ssec:performance} and elaborated above, for the derivation of $N^{*}$ to yield a value $N > 0$ (or $N \ge 1$), we require the term $\frac{\epsilon}{F_{\max} \cdot p_x}$ to be less than 1. 
If $\frac{\epsilon}{F_{\max} \cdot p_x} \geq 1$, i.e., $\epsilon \geq F_{\max} \cdot p_x$, then the marginal gain from the very first computational unit ($N=0$ to $N=1$) is already less than or equal to the threshold $\epsilon$. In such a scenario, $\ln\left(\frac{\epsilon}{F_{\max} \cdot p_x}\right) \geq 0$. Since $\ln(1-p_x)$ is negative, the fraction $\frac{\ln\left(\frac{\epsilon}{F_{\max} \cdot p_x}\right)}{\ln(1-p_x)}$ would be less than or equal to 0. The condition $N > \text{non-positive value}$ would be satisfied by any $N \geq 1$. In this case, $N^{*}$ would effectively be 1 (or even 0, depending on interpretation), meaning no scaling beyond the initial state is justified if one strictly adheres to the $\epsilon$ threshold.

Therefore, the practical application of the $N^{*}$ formula to find a value greater than 1 assumes that the initial potential gain $F_{\max} \cdot p_x$ is greater than the desired negligible gain threshold $\epsilon$. This ensures that there is at least some initial phase where scaling provides a benefit exceeding $\epsilon$.

\section{Additional Experimental Results}
\label{app:results_acc_curve}

\begin{figure}[t]
    \centering
    \subfigure[AIME 2024]{
        \includegraphics[width=0.47\linewidth]{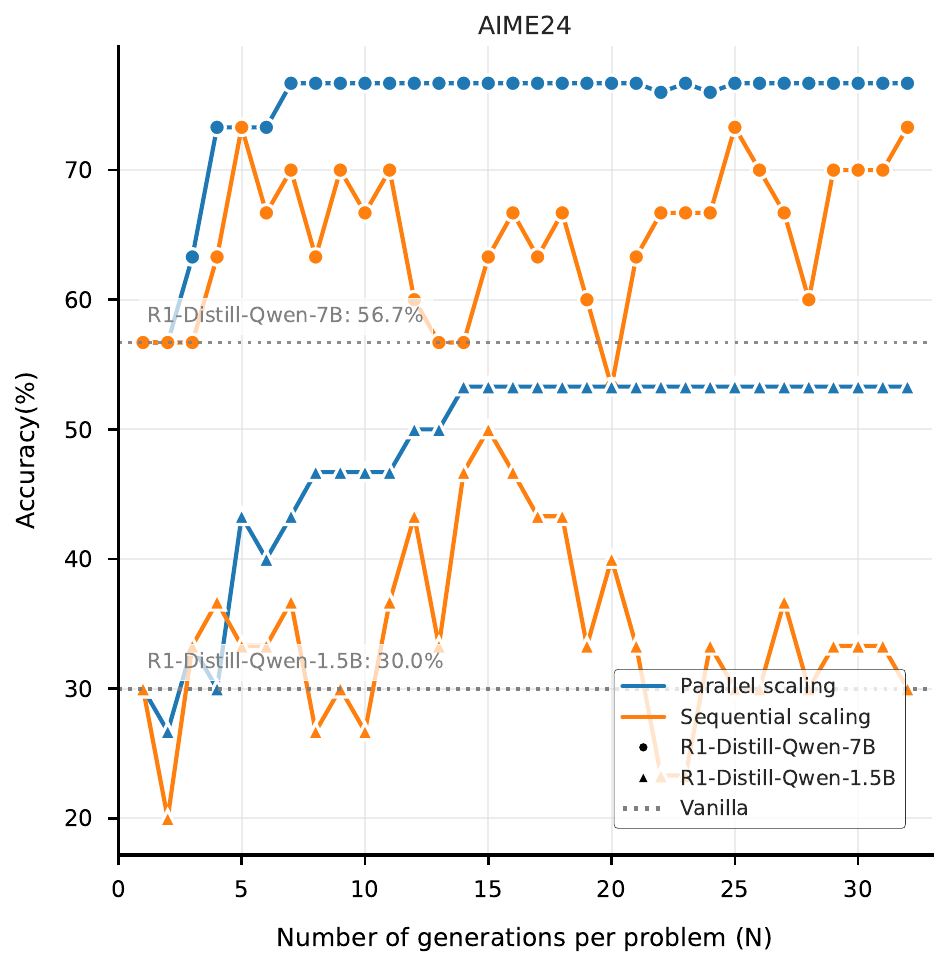}
        \label{fig:acc_aime24}
    }
    \subfigure[AIME 2025]{
        \includegraphics[width=0.47\linewidth]{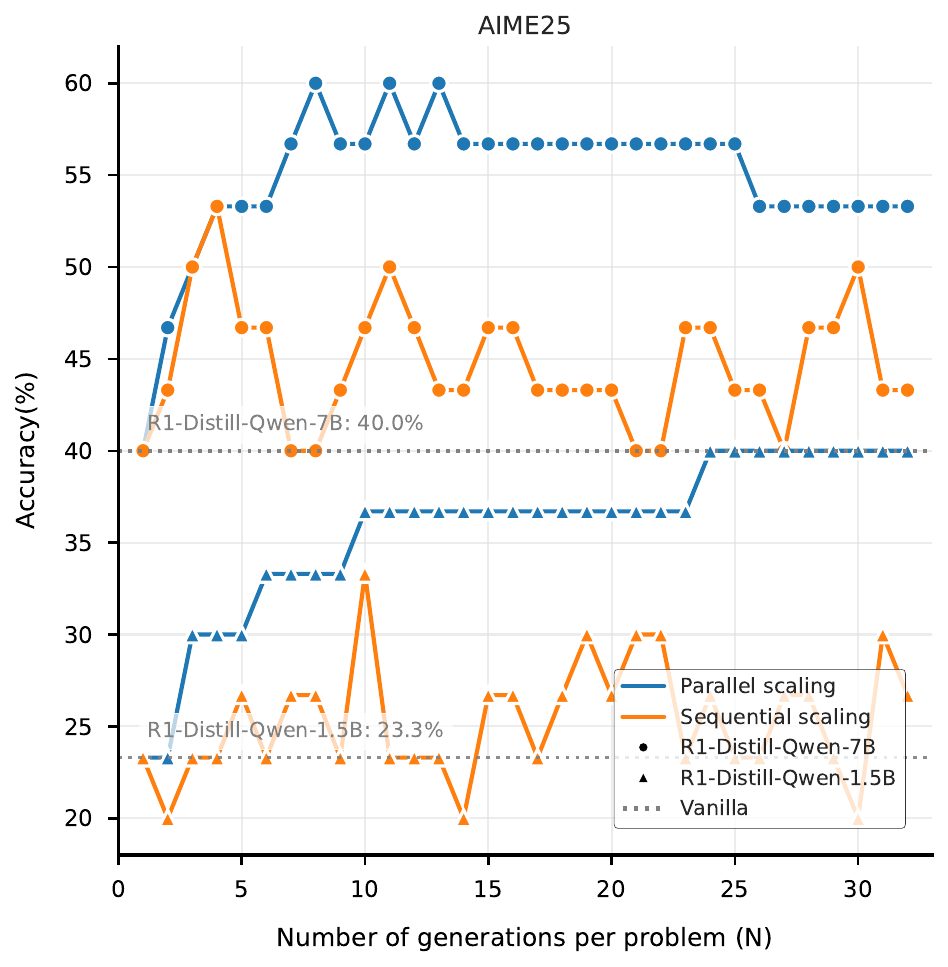}
        \label{fig:acc_aime25}
    }
    \subfigure[MATH-500]{
        \includegraphics[width=0.47\linewidth]{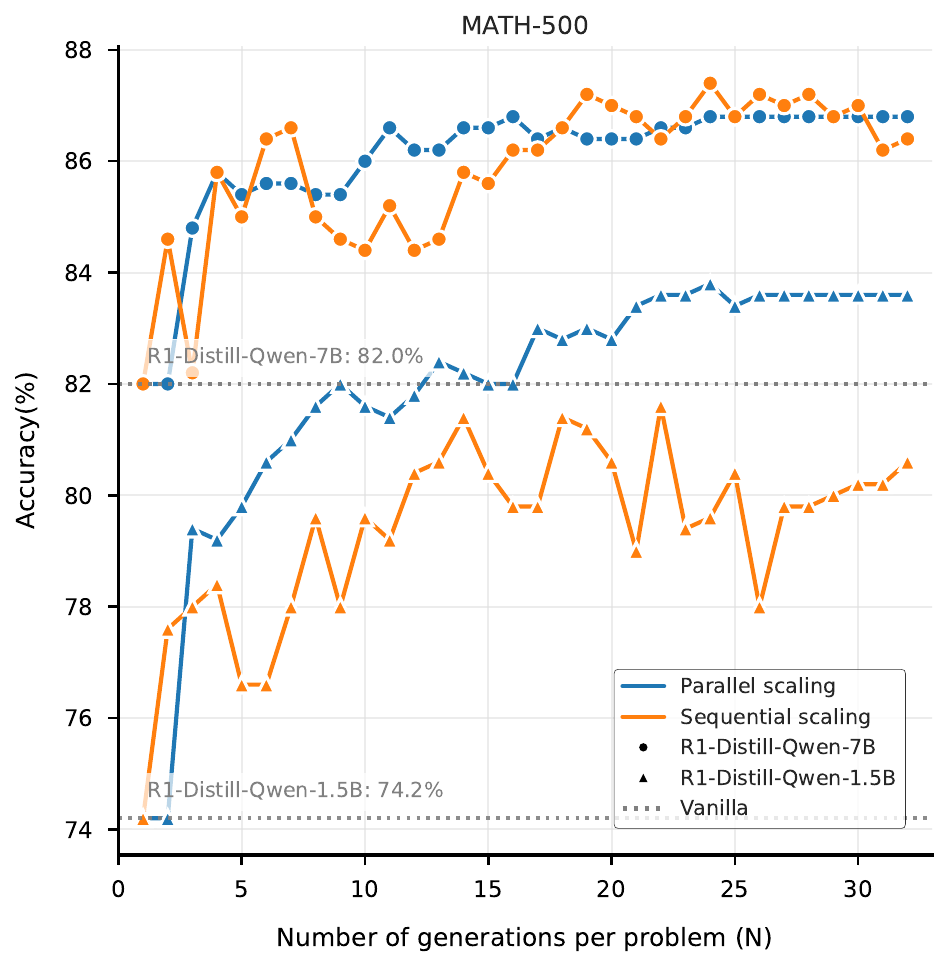}
        \label{fig:acc_math_500}
    }
    \subfigure[GPQA]{
        \includegraphics[width=0.47\linewidth]{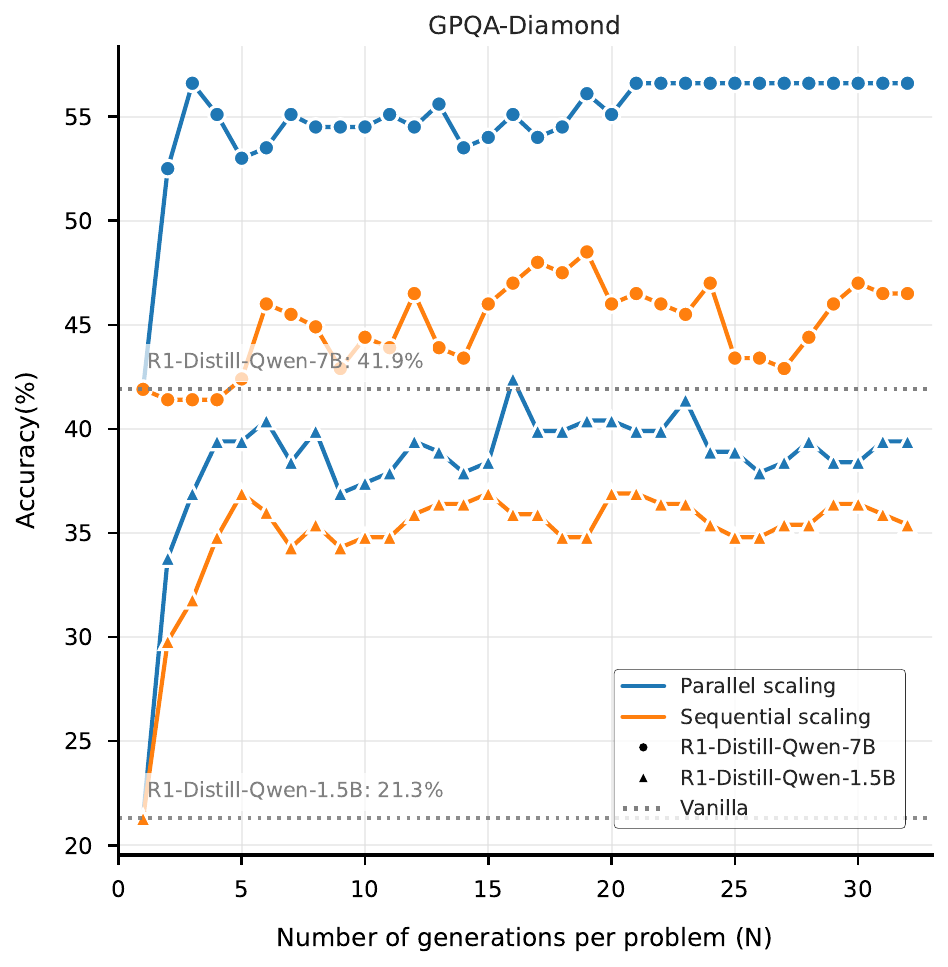}
        \label{fig:acc_gpqa}
    }
    \caption{Accuracy scaling curves across four benchmarks (a) AIME 2024, (b) AIME 2025, (c) MATH-500, and (d) GPQA-Diamond, illustrating how accuracy varies with the number of generations per problem ($N$) under different scaling strategies (parallel vs. sequential) and model sizes.}
    \label{fig:acc_curve}
\end{figure}

Figure~\ref{fig:acc_curve} presents accuracy scaling curves across four benchmarks under different scaling strategies (parallel vs. sequential) and model sizes (1.5B vs. 7B). Across all settings, we observe a consistent trend: initial improvements in accuracy with increased generations per problem ($N$), followed by diminishing returns, forming a clear empirical scaling plateau. 

In addition to the fact that the scaling plateau phenomenon is consistent and universal across tasks, models, and strategies, the above results further lead to several key insights:
1) Test-time scaling enables smaller models to rival or outperform larger, unscaled models. For example, the 1.5B model with sufficient scaling often approaches or exceeds the performance of the 7B vanilla baseline, particularly under parallel scaling. This highlights the practical value of test-time scaling as a lightweight alternative to model scale.
2) Parallel scaling consistently outperforms sequential scaling in terms of final accuracy. While both strategies benefit from increased $N$, parallel scaling achieves better asymptotic performance and exhibits more stable behavior across all benchmarks.
These findings not only align with our theoretical predictions but also suggest that scaling compute at test time, especially via parallel strategies, can be a powerful tool for improving performance without increasing model size.

\end{document}